\documentclass{article}
\usepackage[T1]{fontenc}
\usepackage[utf8]{inputenc}
\usepackage{arxiv}
\usepackage{amsmath,amssymb}
\usepackage{graphicx}
\usepackage{booktabs,multirow,adjustbox}
\usepackage{subcaption}
\usepackage{microtype}
\usepackage{tcolorbox}
\tcbuselibrary{breakable}
\usepackage[ruled,vlined,linesnumbered]{algorithm2e}
\usepackage[numbers,sort&compress]{natbib}
\usepackage[hidelinks]{hyperref}
\renewcommand{\shorttitle}{First Token Matters: Safety Collapse in LRMs}
\hypersetup{pdftitle={First Token Matters: Understanding Safety Collapse in Large Reasoning Models},pdfauthor={Yizheng Yang, Haining Yu, Yuechen Wang, Yikai Hou, Xing Fu, Jinbo Yang, Tianqing Zhu}}
\title{First Token Matters: Understanding Safety Collapse in Large Reasoning Models}
\author{\bfseries Yizheng Yang\textsuperscript{1} \quad Haining Yu\textsuperscript{1}\thanks{Corresponding author: \texttt{yuhaining@hit.edu.cn}.} \quad Yuechen Wang\textsuperscript{1} \quad Yikai Hou\textsuperscript{2} \\
\bfseries Xing Fu\textsuperscript{1} \quad Jinbo Yang\textsuperscript{1} \quad Tianqing Zhu\textsuperscript{3} \\
\normalfont\textsuperscript{1}Harbin Institute of Technology, Harbin, China \\
\normalfont\textsuperscript{2}Guangzhou University, Guangzhou, China \\
\normalfont\textsuperscript{3}City University of Macau, Macau, China}
\date{}

\newtcolorbox{promptbox}[1][]{
  colback=blue!5,
  colframe=blue!50!black,
  title={#1},
  breakable
}

\newtcolorbox{examplebox}[1][]{
  colback=green!5,
  colframe=green!50!black,
  title={#1},
  breakable
}

\begin{document}
\raggedbottom
\maketitle

\begin{abstract}
Large Reasoning Models (LRMs) exhibit strong problem-solving abilities, yet their safety alignment often degrades when handling harmful queries. Existing approaches to improving safety largely rely on additional training or preference optimization, while offering limited understanding of the internal mechanisms behind safety failures. In this work, we investigate this failure through a token-level positional analysis of refusal dynamics and identify a localized vulnerability at the onset of reasoning, which we term Onset Refusal Collapse (ORC). We find that the refusal-related signal of LRMs drops sharply at the first generated token under harmful queries, which is associated with unsafe response generation. Motivated by this finding, we propose SafeToken, a lightweight inference-time intervention that injects a learned continuous safety anchor precisely at reasoning onset. Despite updating only a single token embedding, SafeToken effectively mitigates ORC, improves safety on harmful-query benchmarks, and largely preserves reasoning utility. These results suggest that safety failures in LRMs can arise from a transient breakdown at the critical transition from understanding to generation.

\end{abstract}
\section{Introduction}
LRMs have significantly advanced the state of the art in complex problem-solving by leveraging a multi-step generation process. LRMs rely heavily on Chain-of-Thought (CoT)~\cite{wei2022chain} to handle intricate tasks. By dynamically allocating test-time compute, these models excel in domains requiring rigorous logic, such as mathematics, coding, and scientific reasoning~\cite{lightman2023lets,jaech2024openai,merrill2023expressive}. While this paradigm unlocks unprecedented capabilities, recent studies have revealed that CoT mechanisms in LRMs also introduce safety vulnerabilities~\cite{zhou2025hidden,yong2025self}.

Existing LRM safety methods often rely on additional fine-tuning or preference optimization to align reasoning traces~\cite{rafailov2023direct,Jiang25safechain,wang2025star,jeung2025safepath}. A fundamental limitation of existing studies is that current mitigation methods often treat LRMs as black boxes and rely heavily on human intuition and empirical heuristics to design defense strategies. This highlights a pressing need to uncover the mechanistic cause of safety failures in LRMs.

\begin{figure}[t]
    \centering
    \includegraphics[width=0.95\textwidth]{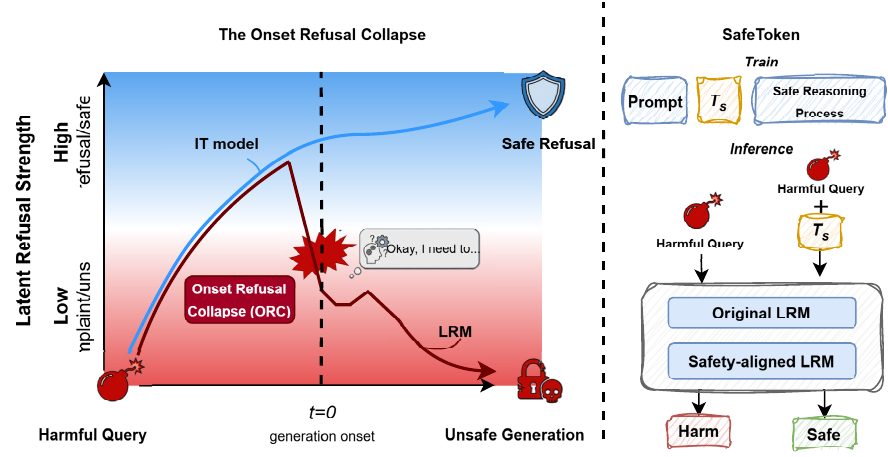}
    \caption{\textbf{Overview} of our work.}
    \label{fig:orc_overview}
\end{figure}
Figure~\ref{fig:orc_overview} illustrates an overview of our work. Building on recent findings that models sharing the same backbone may also share a related safety-relevant internal structure~\cite{DBLP:safetyheads}, we project the latent hidden states of LRMs onto shared refusal vectors extracted from their Instruction-Tuned counterparts. Through this temporal analysis, we discover a critical phenomenon that we term \textbf{Onset Refusal Collapse (ORC)}, as illustrated in Figure~\ref{fig:orc_overview}. Specifically, while the refusal projection strength of LRMs remains aligned with that of IT models during the prompt encoding phase, it drops sharply at the first generated token.

\section{Preliminaries}

\paragraph{Large Reasoning Models.}
We focus on decoder-only Transformer-based LRMs. Given an input query $x$, an LRM generates an intermediate reasoning trajectory $t=(t_1,\dots,t_K)$ before producing the final response $y$:
\begin{equation}
    P(y,t \mid x)=\prod_{i=1}^{K} P(t_i \mid x,t_{<i}) \cdot P(y \mid x,t).
\end{equation}
Unlike standard instruction-tuned models, LRMs rely more heavily on intermediate reasoning traces to solve complex tasks.

\paragraph{Safety-Related Representations.}
Prior mechanistic interpretability work suggests that safety behavior in language models is partly mediated by identifiable internal components, including safety-relevant attention heads~\cite{chen2024finding,DBLP:safetyheads}. ~\cite{DBLP:safetyheads} claim that the pre-training phase shapes fundamental safety capabilities, and that safety heads exhibit substantial overlap across models from the same lineage. Other work shows that refusal behavior can be represented approximately linearly in the residual stream, making it possible to analyze or steer refusal through activation-level interventions~\cite{arditi2024refusal,o2024steering}.

\paragraph{Models and Data.}

We study instruction-tuned (IT) models and LRMs derived from the same pretrained family. For LRMs, we use DeepSeek-R1-Distill-Llama-8B and DeepSeek-R1-Distill-Qwen-7B and 14B~\cite{deepseekai2025deepseekr1incentivizingreasoningcapability}, denoted as R1-8B, R1-7B and R1-14B. As their IT counterparts, we use Llama-3.1-8B-Instruct~\cite{llama3} and Qwen2.5-7B(and 14B)-Instruct~\cite{qwen2,qwen2.5}.\footnote{Actually, R1-7B was distilled from Qwen-2.5-Math-7B-Instruct, but the models ultimately belong to the same base family.} For the datasets, we use AdvBench~\cite{bentov2025universaljailbreaksuffixesstrong}, StrongReject~\cite{souly2024strongreject}, and Alpaca~\cite{alpaca} for analysis.

\section{Locating Safety Collapse in Large Reasoning Models}

In this section, we identify \textbf{Onset Refusal Collapse}, a localized safety failure in LRMs where refusal strength drops sharply at the first generated token. By comparing LRMs with their IT counterparts in a shared refusal subspace, we show that the failure is concentrated at reasoning onset.

\subsection{Safety Failure Induced by Chain of Thought}

We conduct comparative experiments between IT models and LRMs derived from the same backbone. The results suggest that reasoning traces in current LRMs can create a bypass mechanism for safety alignment, thereby motivating a deeper investigation into their internal representations (see Appendix~\ref{app:safety} for more details).

\subsection{The Onset Refusal Collapse of LRMs}

We next ask whether LRMs fail on harmful queries because they genuinely lose refusal capability, or because they fail to preserve refusal-related information that remains encoded in latent space. To answer this, we first construct a shared refusal representation across LRMs and their IT counterparts. Concretely, we identify safety heads in the IT model with EAP-IG~\cite{DBLP:EAP-IG}, and use their layer-wise residual contributions on harmful queries to derive a set of unified refusal vectors, denoted by $\mathcal{V}$ (see Appendix~\ref{app:EAP-IG} and Appendix~\ref{app:refusal_extraction} for details).

\begin{table}[t]
\caption{\textbf{Attack Success Rate (ASR) on JailbreakBench-Behavior under steering with refusal vectors.} NS denotes Negative Steering, and PS denotes Positive Steering.}
\label{homogeneity}
\centering
\small
\setlength{\tabcolsep}{8pt}
\renewcommand{\arraystretch}{1.05}
\begin{tabular}{lccc}
\toprule
\textbf{Model} & \textbf{Before} & \textbf{NS} & \textbf{PS} \\
\midrule
R1-8B & 45\% & 75\% & 21\% \\
R1-7B & 59\% & 87\% & 27\% \\
\bottomrule
\end{tabular}
\end{table}

We test the transferability of $\mathcal{V}$ through activation steering on R1-7B and R1-8B. As shown in Table~\ref{homogeneity}, steering along these vectors yields a clear two-sided effect on model responses. This result suggests that LRMs and their IT counterparts derived from the same backbone share similar latent directions associated with refusal. Based on this shared representational structure, we next examine how these directions behave across token positions during generation.

We project the hidden states of each generated token onto $\mathcal{V}$ and track the resulting \textit{Refusal Projection Strength}:
\begin{equation}
    S_t = \frac{1}{L} \sum_{l=1}^{L} (h_{l,t} \cdot v_l)
\end{equation}
where $L$ is the total number of layers and higher $S_t$ indicates stronger refusal-related activation. By comparing the evolution of $S_t$ across token positions, we can examine how refusal-related representations change over the course of generation.

\begin{figure}[tb]
    \centering
    \begin{subfigure}[b]{0.48\linewidth}
        \centering
        \includegraphics[width=\linewidth]{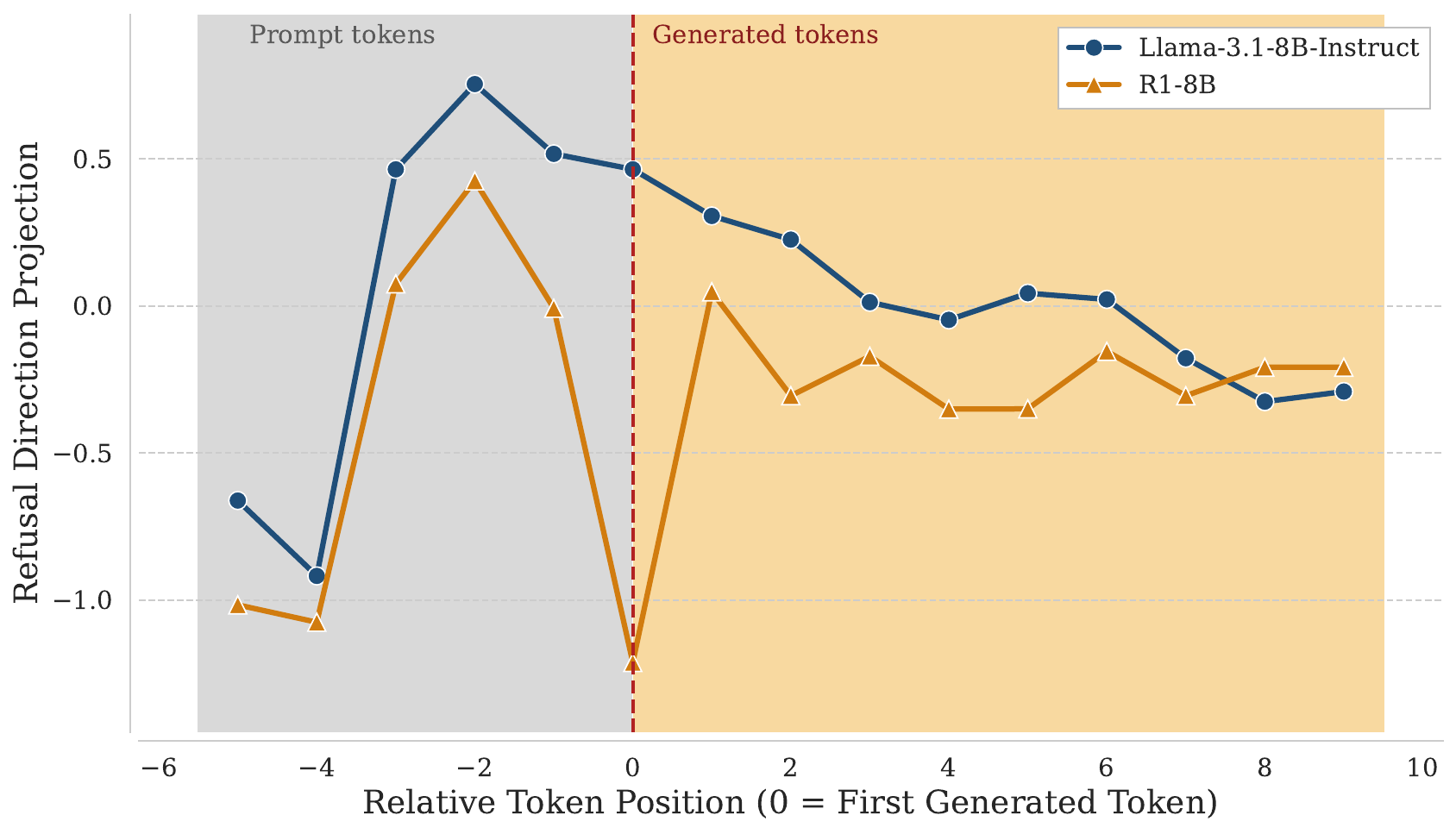}
        \caption{Trajectory comparison for Llama.}
        \label{fig:trajectory_llama}
    \end{subfigure}
    \hfill
    \begin{subfigure}[b]{0.48\linewidth}
        \centering
        \includegraphics[width=\linewidth]{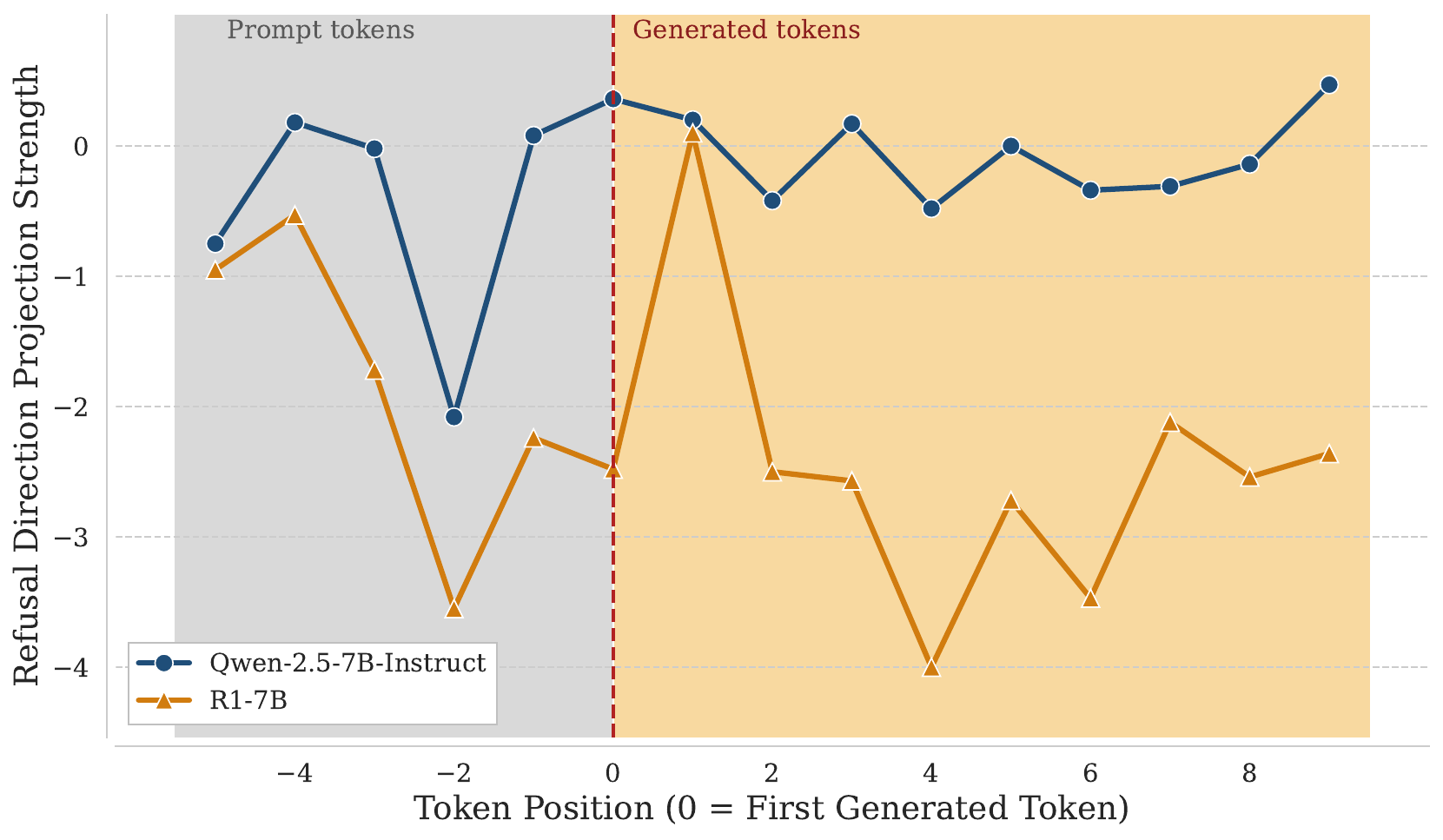}
        \caption{Trajectory comparison for Qwen.}
        \label{fig:trajectory_qwen}
    \end{subfigure}

    \caption{\textbf{Trajectory comparison analysis.}
    The visualization illustrates the refusal-projection trajectories of LRMs and their IT counterparts on harmful prompts.}
    \label{fig:ORC}
\end{figure}

Figure~\ref{fig:ORC} shows that, under harmful queries, LRMs remain partially aligned with their IT counterparts during prompt encoding ($t<0$), suggesting that they may still retain latent awareness of risk. However, a sharp divergence emerges exactly at the first generated token ($t=0$): while the IT model maintains strong refusal activation, the LRM exhibits a sudden drop toward the non-refusal region. We term this abrupt decline \textbf{Onset Refusal Collapse (ORC)}. This localization identifies generation onset as a possible failure point. Results for harmless inputs are provided in Appendix~\ref{app:benign}.

To quantify this effect beyond single trajectories, we measure ORC relative to each matched IT/LRM pair. Specifically, we compute the matched-pair refusal gap $G_t = S_t^{IT} - S_t^{LRM}$ and define the normalized onset change as
\begin{equation}
R_{\mathrm{ORC}} =
\frac{G_0 - G_{-1}}
{\frac{1}{4}\left(|S_{-1}^{IT}| + |S_{-1}^{LRM}| + |S_0^{IT}| + |S_0^{LRM}|\right)}.
\end{equation}
Positive values indicate that the IT/LRM refusal gap widens at generation onset. In Table~\ref{tab:orc_statistics}, \textbf{Harmful $n$} denotes the number of harmful prompts used for estimation; \textbf{Mean $R_{\mathrm{ORC}}$} reports the average normalized onset-induced change in the matched IT/LRM refusal gap; and \textbf{95\% CI} denotes the bootstrap confidence interval of the mean. Results for harmless inputs are provided in Appendix~\ref{app:benign}.
\begin{table}[tb]
\caption{\textbf{Aggregate ORC statistics on harmful prompts.}
$R_{\mathrm{ORC}}$ measures the normalized onset-induced change in the matched IT/LRM refusal gap.}
\label{tab:orc_statistics}
\centering
\small
\setlength{\tabcolsep}{4pt}
\renewcommand{\arraystretch}{1.08}
\begin{tabular}{lccc}
\toprule
\textbf{Matched Pair} & \textbf{Harmful $n$} & \textbf{Mean $R_{\mathrm{ORC}}$} & \textbf{95\% CI} \\
\midrule
Llama-3.1-8B-Instruct vs R1-8B & 100 & 189.41\% & [179.82\%, 199.81\%] \\
Qwen2.5-7B-Instruct vs R1-7B & 100 & 20.00\% & [14.73\%, 25.24\%] \\
Qwen2.5-14B-Instruct vs R1-14B & 100 & 100.11\% & [91.37\%, 108.57\%] \\
\bottomrule
\end{tabular}
\end{table}
\subsection{Linking ORC to Safety Failure}

\begin{table}[tb]
\caption{ASR on the StrongReject benchmark under onset interventions.}
\label{tab:prefix_length}
\centering
\small
\setlength{\tabcolsep}{8pt}
\renewcommand{\arraystretch}{1.1}
\begin{tabular}{lcc}
\toprule
\textbf{Onset Intervention} & \textbf{R1-7B} & \textbf{R1-8B} \\
\midrule
0 (Vanilla LRM)             & 49.20 & 37.38 \\
Prefix Length = 1           & 20.13 & 4.47 \\
Prefix Length = 2           & 22.68 & 3.83 \\
Prefix Length = 3           & 25.88 & \textbf{3.51} \\
Prefix Length = 4           & 27.80 & 8.31 \\
Prefix Length = 5           & 15.65 & 7.35 \\
Prefix Length = 6           & 25.88 & 9.90 \\
Prefix Length = 7           & 19.81 & 7.67 \\
Prefix Length = 8           & 24.60 & 10.54 \\
\midrule
Refusal Direction Injection & \textbf{15.02} & 7.35 \\
\bottomrule
\end{tabular}
\end{table}

A central question is whether ORC is directly tied to the safety failure of LRMs on harmful queries. To answer it more directly, we intervene at the exact onset position where ORC occurs and examine the resulting change in safety performance.

We first conduct extensive sampling of responses to harmful queries from R1-8B and R1-7B. We find that the vast majority of responses begin with affirmative tokens such as ``Okay'' or ``Alright'' even in cases where the model eventually refuses. This suggests that the collapse at reasoning onset may play a functional role in shaping the model's downstream safety behavior.

To test this connection, we evaluate R1-7B and R1-8B on StrongReject by force-decoding safety-related prefixes of lengths ranging from 1 to 8 tokens at the first generation step. Examples include a single-token prefix such as ``Safety'' and the full eight-token phrase ``Let's think about safety first.'' used in SafePath~\cite{jeung2025safepath}. We additionally evaluate a direct intervention that injects the refusal direction at the first generated token. The detailed setup is provided in Appendix~\ref{app:prefixes}.

As shown in Table~\ref{tab:prefix_length}, onset interventions at the first generated token consistently reduce ASR relative to the vanilla LRM across both models. Even single-token safety prefixes yield large gains, indicating that safety performance is highly sensitive to the initialization state. This view is further supported by refusal direction injection, which achieves similar improvements. These results suggest that ORC is not only temporally localized, but also makes an important causal contribution to unsafe behavior in LRMs on harmful queries.

We further investigate the possible source of ORC and observe that this onset pattern is associated with a strong tendency to initiate reasoning with affirmative response-formatting tokens. We defer the detailed discussion and supporting evidence to Appendix~\ref{app:source}.

\section{SafeToken: Anchoring Safety at Reasoning Onset}

\subsection{Methodology}

\subsubsection{Motivation}

Since ORC is sharply localized at the first generated token, an effective intervention should act precisely at reasoning onset and directly reshape the model's initialization state before unsafe reasoning unfolds. As shown in Table~\ref{tab:prefix_length}, safety performance is highly sensitive to the specific content injected at this position, and shorter safety-related prefixes can even outperform longer ones. This suggests that the key factor is not the use of a fixed natural-language statement itself, but whether the injected signal can effectively mitigate the ORC phenomenon. However, searching for such an intervention in the discrete space of natural-language tokens is inherently limited by the tokenizer vocabulary and may easily interfere with utility. Therefore, inspired by prompt tuning~\cite{zhang2024safetypromptfinetuning}, we propose to learn a single optimized continuous representation that serves as a safety anchor for LRMs at reasoning onset.

\subsubsection{Mechanism of SafeToken}

We introduce a continuous ``Soft Token'' optimized specifically for safety. By directly learning the optimal latent embedding to be injected at the onset, we can effectively mitigate the ORC phenomenon.

\paragraph{Soft Token Optimization.}
We expand the tokenizer's vocabulary to include the special control token $T_{s}$ and explicitly place it exactly at the generation onset ($t=0$). During the optimization process, we strictly freeze all parameters of the base LRM and update the embedding vector of $T_{s}$ via the standard causal language modeling loss.

\paragraph{Inference-Time Injection.}
Once trained, the soft token serves as a plug-and-play safety anchor. During inference, given any user query $x$, we deterministically append the learned $T_{s}$ to the end of the input sequence. By occupying the critical first position of the output sequence, $T_{s}$ mitigates ORC and guides the subsequent reasoning trajectory. The generation probability is thus modified from the standard formulation to explicitly condition on this injected token:
\begin{equation}
    P(y, t | x) = \prod_{i=1}^{K} P(t_i | x \oplus T_{s}, t_{<i}) \cdot P(y | x \oplus T_{s}, t)
\end{equation}
where $\oplus$ denotes sequence concatenation, $t=(t_1, \dots, t_K)$ represents the intermediate reasoning trace, and $y$ denotes the final response.

\begin{table*}[!t]
\caption{\textbf{Evaluation results on safety, jailbreak, over-refusal, and utility benchmarks.}
Lower is better for safety-related metrics ($\downarrow$), and higher is better for utility metrics ($\uparrow$).
SR, Adv., WJ, XS, SC, SP, and ST denote StrongReject, AdvBench, WildJailbreak, XSTest, SafeChain, SafePath, and SafeToken, respectively.}
\label{tab:main_results}
\centering
\small
\setlength{\tabcolsep}{2.0pt}
\renewcommand{\arraystretch}{1.03}
\begin{tabular}{lcccccccccc}
\toprule
\textbf{Method}
& \textbf{Par.}
& \textbf{SR}
& \textbf{Adv.}
& \textbf{GCG}
& \textbf{WJ}
& \textbf{XS}
& \textbf{MMLU}
& \textbf{MATH}
& \textbf{AIME}
& \textbf{GPQA} \\
\midrule

\multicolumn{11}{c}{\textbf{R1-8B}} \\
\midrule
Orig.  & -- & 37.3 & 35.0 & 62.0 & 50.8 & \underline{5.2} & \textbf{69.9} & \underline{75.2} & \textbf{40.0} & \textbf{46.0} \\
SC     & 8B & 33.6 & 22.3 & 43.5 & 37.3 & \textbf{3.6} & \underline{66.5} & 72.6 & 30.0 & 42.4 \\
Pref.  & -- & 10.5 & 9.0  & 25.3 & 29.8 & 22.4 & 63.0 & 67.4 & \underline{33.3} & 41.4 \\
SP     & 8B & 2.6  & 8.1  & \underline{6.5} & \textbf{0.8} & 74.8 & 58.2 & 72.8 & 23.3 & \underline{43.4} \\
STAR   & 8B & \textbf{0.3} & \textbf{0.0} & \textbf{3.8} & \underline{7.8} & 25.2 & 65.0 & \textbf{76.6} & 30.0 & \underline{43.4} \\
ST     & \textbf{4096} & \underline{2.2} & \underline{0.6} & 25.3 & 23.2 & 11.6 & 66.0 & \underline{75.2} & 26.7 & 42.9 \\
\midrule

\multicolumn{11}{c}{\textbf{R1-7B}} \\
\midrule
Orig.  & -- & 49.2 & 50.3 & 64.5 & 52.7 & \underline{13.2} & \textbf{63.5} & 78.4 & \underline{46.7} & \textbf{52.0} \\
SC     & 7B & 41.2 & 27.0 & 51.3 & 39.5 & \textbf{12.8} & \underline{61.8} & 80.0 & \underline{46.7} & 47.0 \\
Pref.  & -- & 24.6 & 14.6 & 18.8 & 35.8 & 38.0 & 60.3 & 77.4 & \underline{46.7} & 39.9 \\
SP     & 7B & 6.4  & 2.7  & \underline{13.0} & \underline{17.6} & 54.8 & 58.2 & 77.4 & 40.0 & \underline{51.5} \\
STAR   & 7B & \textbf{0.2} & \textbf{0.0} & \textbf{10.8} & \textbf{15.1} & 41.6 & 61.1 & \underline{81.2} & 36.7 & 45.5 \\
ST     & \textbf{3584} & \underline{4.8} & \underline{2.5} & 16.0 & 24.9 & 37.2 & 61.6 & \textbf{82.0} & \textbf{53.3} & 51.0 \\
\midrule

\multicolumn{11}{c}{\textbf{Qwen3-8B}} \\
\midrule
Orig.  & -- & 5.8 & 1.2 & 43.5 & 39.8 & \textbf{1.6} & \underline{82.8} & 79.6 & 40.0 & 53.5 \\
SC     & 8B & 38.3 & 28.7 & 54.5 & 40.1 & \underline{4.8} & 81.8 & 81.8 & 30.0 & 57.6 \\
Pref.  & -- & \underline{0.3} & \underline{0.4} & 11.0 & 18.4 & 6.0 & 82.0 & 82.4 & \textbf{53.3} & \textbf{61.1} \\
SP     & 8B & 1.9 & 1.4 & \underline{6.8} & \underline{14.7} & 9.2 & \textbf{83.1} & 78.2 & 43.3 & \underline{60.1} \\
STAR   & 8B & \textbf{0.0} & \textbf{0.0} & \textbf{3.4} & \textbf{5.5} & 44.8 & 73.1 & \textbf{83.6} & \underline{46.7} & 42.9 \\
ST     & \textbf{4096} & \textbf{0.0} & \textbf{0.0} & 8.3 & 16.4 & 11.2 & 78.5 & \underline{82.8} & 43.3 & 46.0 \\
\bottomrule
\end{tabular}
\end{table*}

\subsection{Experiments}

\subsubsection{Experiment Setup}

\paragraph{Baselines.} We select R1-8B and R1-7B as the target LRMs. To further examine the transferability, we include Qwen3-8B~\cite{qwen3technicalreport} in the experiments. We compare SafeToken with SafeChain~\cite{Jiang25safechain}, SafePath~\cite{jeung2025safepath}, STAR-1~\cite{wang2025star}, and Safety Prefix, which is an inference-time variant of SafePath. More details are provided in Appendix~\ref{app:setup}.

\paragraph{SafeToken Setup.} For fair comparison, we construct our training data following the setup of STAR-1 and use 1,000 samples to optimize the proposed continuous safety anchor. The learnable token is initialized with the embedding of the token ``Safety''. The token embedding is trained using the standard causal language modeling loss with a batch size of 8. We train for 15 epochs on R1-8B and 20 epochs on R1-7B, utilizing the AdamW optimizer with a learning rate of $5 \times 10^{-3}$.

\paragraph{Evaluation.} We evaluate performance across four dimensions: safety, jailbreak robustness, over-refusal, and utility. For safety, we employ StrongReject~\cite{souly2024strongreject} and AdvBench~\cite{zou2023universal}, utilizing Llama-Guard-3-8B~\cite{dubey2024llama} as an automated evaluator to classify the safety of the generated responses. To assess robustness against jailbreak attacks, we utilize GCG~\cite{bentov2025universaljailbreaksuffixesstrong} using HarmBench~\cite{mazeika2024harmbench} and WildJailbreak~\cite{wildteaming2024}. Furthermore, we measure over-refusal using the XSTest dataset~\cite{rottger2023xstest}, employing DeepSeek-Chat~\cite{deepseekai2025deepseekr1incentivizingreasoningcapability} to evaluate whether the model improperly refuses to answer benign queries or comply with harmful queries. For utility, we report results on MMLU~\cite{hendryckstest2021mmlu}, MATH500~\cite{lightman2023lets}, AIME2024~\cite{maa_aime} and GPQA-Diamond~\cite{rein2024gpqa}.

\subsubsection{Main Results}

We conducted a comparative analysis of different defense mechanisms, evaluating their performance across both safety and utility benchmarks. The quantitative results are presented in Table~\ref{tab:main_results}.

\paragraph{Safety and Utility Trade-off.}
SafeToken achieves strong safety performance on several harmful-query benchmarks while requiring only a single token embedding to be optimized. At the same time, its utility remains competitive. Overall, SafeToken provides a favorable trade-off between safety, utility, and training cost. SafeToken shows limited robustness under strong jailbreak attacks. A likely reason is that jailbreak prompts often act earlier in the pipeline by altering the model's internal assessment of whether a query is harmful. We further discuss the safety performance in Appendix~\ref{app:performance}.

\paragraph{Training Efficiency.}
SafeToken requires updating only a single token embedding, whose size equals the model's hidden dimension. Consequently, the entire optimization process converges within minutes on a single GPU.

\begin{figure}[htbp]
    \centering
    \includegraphics[width=0.7\linewidth]{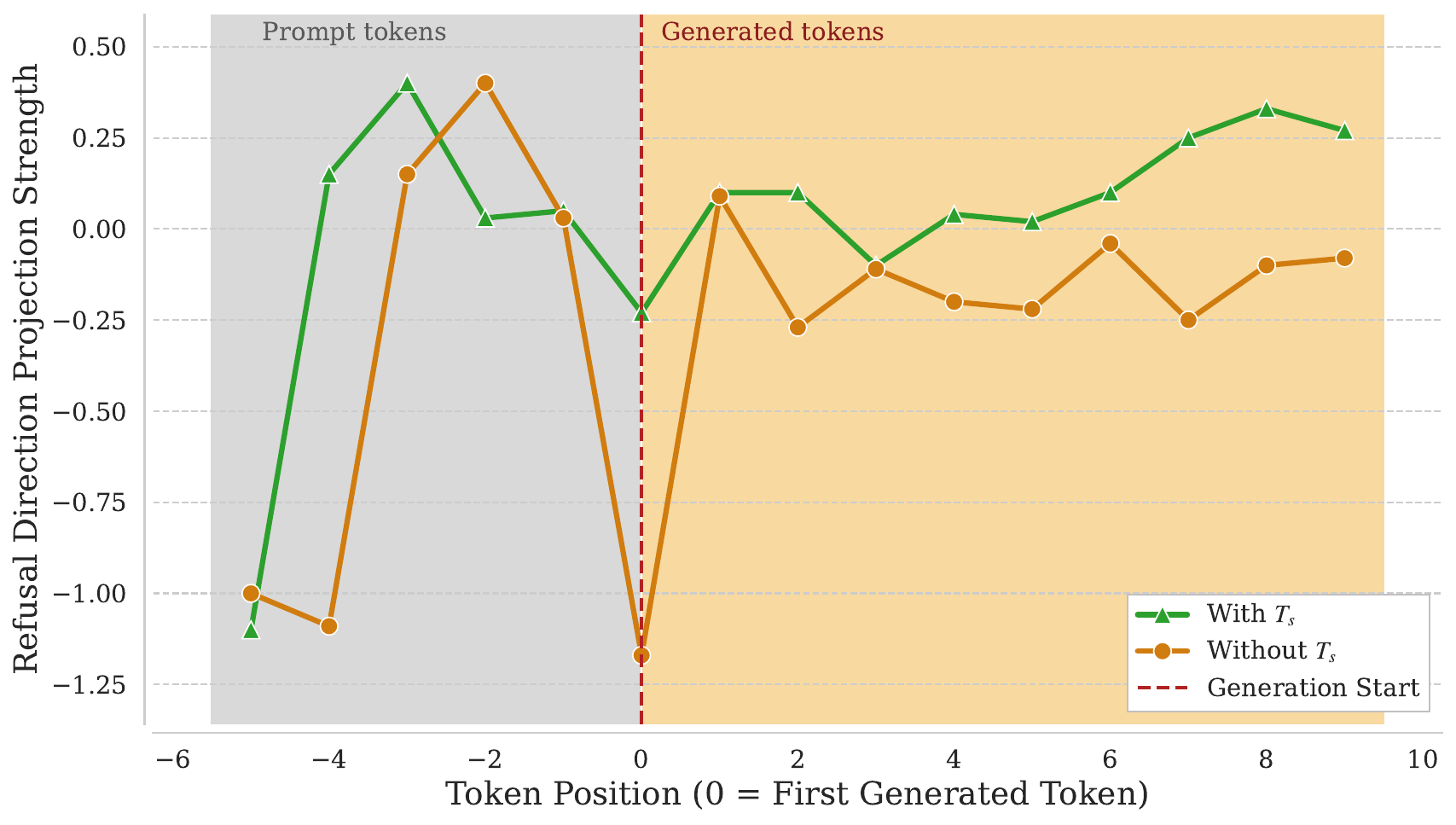}
    \caption{SafeToken (With $T_s$) successfully prevents the sharp decline (ORC) observed in R1-8B (Without $T_s$) precisely at $t=0$.}
    \label{fig:orc_mitigation}
\end{figure}

\paragraph{Mitigating the ORC Phenomenon.}
As established previously, mitigating the ORC phenomenon directly improves the safety performance of LRMs. Figure~\ref{fig:orc_mitigation} visually demonstrates the mechanistic impact of SafeToken. By explicitly anchoring the learned safety representation at the end of the prompt, the refusal projection strength remains robust during the critical initialization phase. Consequently, avoiding this initial collapse leads to a significantly higher refusal strength throughout the subsequent reasoning trajectory.

\begin{table}[!t]
\caption{\textbf{Effect of adding SafeToken to SafePath and STAR-1.} Lower is better for safety metrics.}
\label{tab:combine_safetoken}
\centering
\small
\setlength{\tabcolsep}{4pt}
\renewcommand{\arraystretch}{1.08}
\begin{tabular}{lllcc}
\toprule
\textbf{Model} & \textbf{Method} & \textbf{Metric} & \textbf{Before} & \textbf{+ SafeToken} \\
\midrule

\multirow{6}{*}{\textbf{R1-7B}}
& \multirow{3}{*}{SafePath}
& StrongReject   & 6.4  & \textbf{2.6} \\
& & AdvBench       & 2.7  & \textbf{0.8} \\
& & WildJailbreak  & 17.6 & \textbf{13.2} \\
\cmidrule(lr){2-5}
& \multirow{3}{*}{STAR-1}
& StrongReject    & 0.2  & \textbf{0.0} \\
& & AdvBench       & \textbf{0.0} & \textbf{0.0} \\
& & WildJailbreak  & 15.1 & \textbf{12.0} \\
\midrule

\multirow{6}{*}{\textbf{R1-8B}}
& \multirow{3}{*}{SafePath}
& StrongReject    & \textbf{2.6} & 4.2 \\
& & AdvBench       & 8.1  & \textbf{0.8} \\
& & WildJailbreak  & \textbf{0.8} & 11.3 \\
\cmidrule(lr){2-5}
& \multirow{3}{*}{STAR-1}
& StrongReject    & \textbf{0.3} & \textbf{0.3} \\
& & AdvBench       & \textbf{0.0} & \textbf{0.0} \\
& & WildJailbreak  & \textbf{7.8} & 8.2 \\
\midrule

\multirow{6}{*}{\textbf{Qwen3-8B}}
& \multirow{3}{*}{SafePath}
& StrongReject    & 1.9  & \textbf{0.0} \\
& & AdvBench       & 1.4  & \textbf{0.0} \\
& & WildJailbreak  & 14.7 & \textbf{9.1} \\
\cmidrule(lr){2-5}
& \multirow{3}{*}{STAR-1}
& StrongReject    & \textbf{0.0} & \textbf{0.0} \\
& & AdvBench       & \textbf{0.0} & \textbf{0.0} \\
& & WildJailbreak  & 5.5  & \textbf{4.9} \\
\bottomrule

\end{tabular}
\end{table}

\paragraph{Compatibility with Existing Safety Methods.}
To further evaluate the modularity of SafeToken, we combine it with two representative defenses, SafePath and STAR-1, and compare the resulting safety performance before and after adding SafeToken. As shown in Table~\ref{tab:combine_safetoken}, SafeToken improves or preserves safety performance across most settings. These results suggest that SafeToken is not only effective as a standalone method, but can also serve as a complementary safety component for existing reasoning-time defenses. We provide more details about utility performance in Appendix~\ref{app:combined}.

\begin{table}[!t]
\caption{Impact of injection position on safety benchmarks. Lower ASR ($\downarrow$) indicates better safety performance.}
\label{tab:injection_position}
\centering
\small
\setlength{\tabcolsep}{6pt}
\renewcommand{\arraystretch}{1.08}
\begin{tabular}{llcc}
\toprule
\textbf{Model} & \textbf{Position} & \textbf{StrongReject} ($\downarrow$) & \textbf{AdvBench} ($\downarrow$) \\
\midrule
\multirow{2}{*}{\textbf{R1-8B}}
& Front Injection & 41.9 & 37.9 \\
& \textbf{SafeToken} & \textbf{2.2} & \textbf{0.6} \\
\midrule
\multirow{2}{*}{\textbf{R1-7B}}
& Front Injection & 64.9 & 56.4 \\
& \textbf{SafeToken} & \textbf{4.8} & \textbf{2.5} \\
\bottomrule
\end{tabular}
\end{table}

\subsubsection{Sensitivity Analysis: SafeToken Position}

To validate the necessity of anchoring position, we compare our proposed onset injection with the conventional front-injection baseline used in most prompt-tuning studies~\cite{lester2021power}. For the Front Injection setup, we prepend the soft token $T_s$ to the very beginning of the user prompt to act as an implicit system prompt. For a fair comparison, we train the front-injected token with the same dataset, configuration, and hyperparameters used for SafeToken. We then evaluate both configurations on the StrongReject and AdvBench benchmarks using the R1-8B and R1-7B models.

As demonstrated in Table \ref{tab:injection_position}, the Front Injection strategy results in a significant degradation of safety performance compared to SafeToken, indicating that the position of SafeToken is critical.

\section{Related Work}

\paragraph{Safety in LRMs.} \cite{zhou2025hidden} find that stronger reasoning ability leads to greater potential harm. This trade-off is exemplified by self-jailbreaking~\cite{yong2025self,mao2025models}, a phenomenon where models exploit their own reasoning processes to rationalize complying with malicious queries. Furthermore, adversaries can weaponize this mechanism through CoT-based attacks~\cite{kuo2025h,zhao2025chain}. To counter the unique vulnerabilities introduced by complex reasoning traces, recent studies have explored several defense paradigms~\cite{Jiang25safechain,jeung2025safepath,han2025safeswitch,ghosal2026safety,peng2025large}. \cite{yin2025refusal,huang-etal-2025-path,wu2025read} also discover other forms of safety collapse in LRMs.

\paragraph{Impact of Tokens.} Prompt tuning learns continuous input embeddings while keeping the backbone model frozen~\cite{lester2021power}. Recent safety methods build on this idea to reduce unsafe generations without full-parameter fine-tuning~\cite{zhang2024safetypromptfinetuning,yuan2025promptguard}. Meanwhile, prior work shows that the effect of prompt-based control depends not only on the prompt representation itself, but also on where it is injected in the model~\cite{whendoprompting,DBLP:impactofposition,yang-etal-2025-position}.

\section{Conclusion}

In this work, we investigate why safety alignment in LRMs can fail under harmful queries. Through a token-level analysis of refusal dynamics, we identify a localized vulnerability at the onset of reasoning. Motivated by this observation, we propose SafeToken, a lightweight inference-time intervention that learns a continuous safety anchor and injects it precisely at reasoning onset. SafeToken effectively mitigates this collapse and improves safety performance on harmful-query benchmarks while preserving general reasoning utility.

Overall, our results highlight reasoning onset as a critical stage for safety in LRMs and provide evidence that lightweight, mechanism-driven interventions can effectively improve safety behavior.

\section{Limitations}

Our study focuses on a specific safety failure mode in LRMs, namely the collapse of refusal behavior at reasoning onset. Although SafeToken improves safety across multiple settings, its gains are less pronounced in some stronger attack scenarios. One possible reason is that such attacks may interfere with the model's recognition of harmfulness itself, whereas SafeToken mainly targets cases where risk awareness is already present but fails to be translated into refusal behavior due to ORC. A possible direction for future work is to integrate SafeToken into the training pipeline. Such a combination may improve robustness against attacks that act earlier in the safety decision process.

\section*{Acknowledgments}

We thank anonymous reviewers for valuable comments. This work was supported in part by the National Natural Science Foundation of China under Grant 62302122, National Key Research and Development Program of China  under Grant 2025YFB3109803 and Heilongjiang Provincial Natural Science Foundation of China under Grant JQ2024F001.

\clearpage
\bibliographystyle{plainnat}
\bibliography{reference}

\clearpage
\appendix
\section{Safety Performance} \label{app:safety}

Recent studies have highlighted a critical trade-off in LRMs: while CoT enhances performance on complex tasks, it also places substantial pressure on safety mechanisms.

\begin{figure}[htbp]
    \centering
    \includegraphics[width=1.0\textwidth]{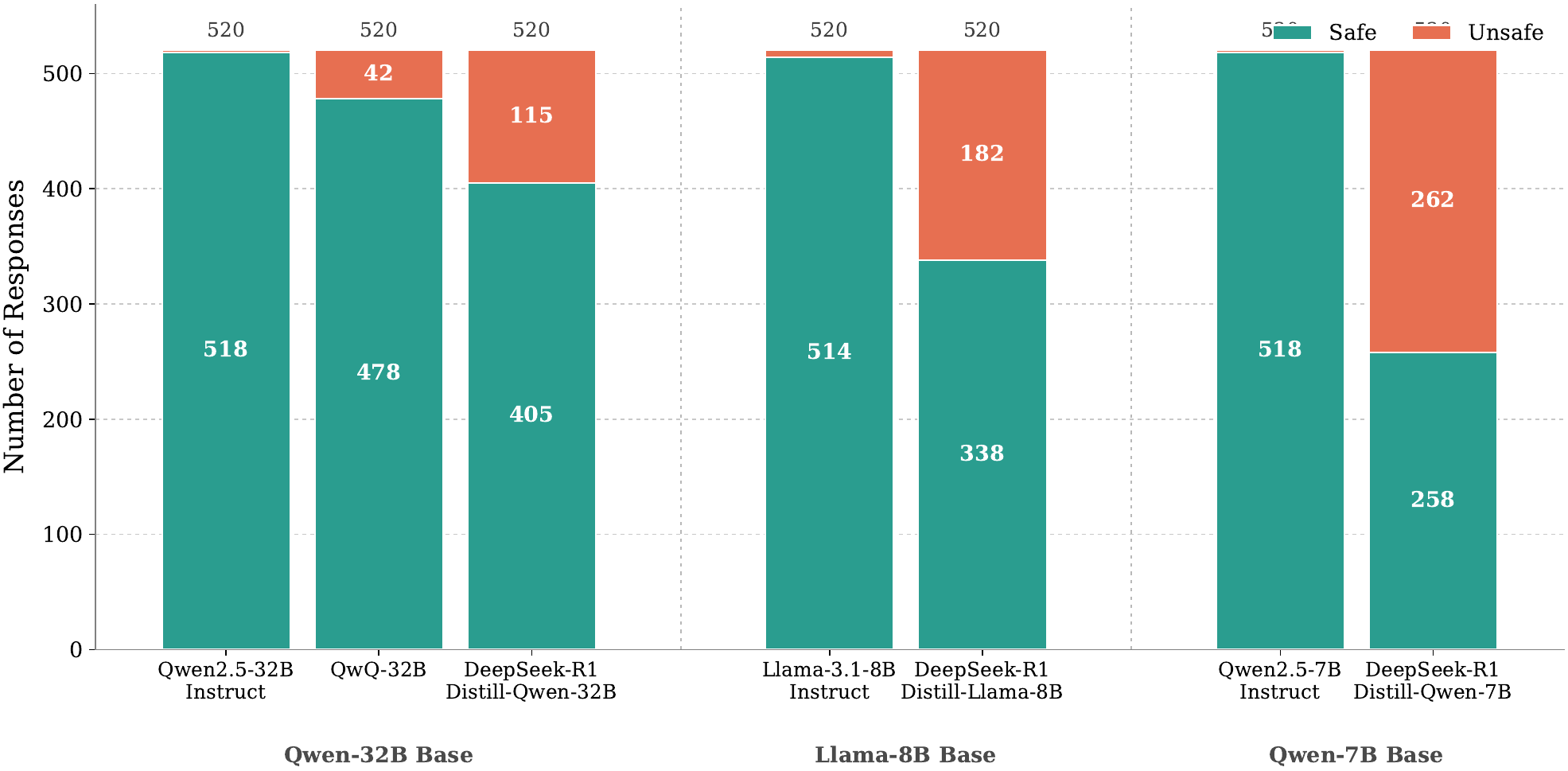}
    \caption{Safety performance comparison between IT baselines and derived LRMs on AdvBench.}
    \label{fig:LRM_and_IT_ASR}
\end{figure}

To verify this observation, we conduct comparative experiments between IT models and LRMs derived from the same backbone. As illustrated in Figure~\ref{fig:LRM_and_IT_ASR}, the IT models maintain robust refusal rates on AdvBench. In contrast, the corresponding LRMs exhibit a marked increase in unsafe responses.

\section{Discovery of Safety Heads} \label{app:EAP-IG}

\subsection{EAP-IG Introduction and Data Preparation}
EAP-IG identifies critical model components by comparing attribution patterns between \emph{counterfactual input pairs}. Intuitively, the method requires two closely matched inputs that differ only in the target behavior, so that the resulting attribution difference can be used to isolate components that are causally related to the behavior of interest. In our setting, this means constructing paired prompts that preserve the same semantic content while differing in whether they elicit a safe refusal or an unsafe response onset.

However, applying this algorithm to LRMs is particularly challenging. In practice, LRMs tend to begin both safe and unsafe responses with the same or highly similar onset tokens. As a result, the behavioral divergence of interest is often not reflected in the surface form of the first generated token, even though the underlying safety state may already differ internally. This makes it difficult to construct clean counterfactual pairs for EAP-IG.

To support this analysis, we manually design a prompt construction procedure to generate such counterfactual pairs. Specifically, we use a locally deployed LLM to rewrite harmful prompts into closely matched safe counterparts. As shown in the prompt box below, the rewriting prompt explicitly instructs the model to preserve the original wording, structure, and style as much as possible, while modifying only the minimum content necessary to remove harmful intent. This controlled setup allows EAP-IG to more reliably identify the components that contribute to the behavioral difference, rather than capturing unrelated variations in content or style.

We provide two representative examples of this rewriting process in the example boxes below. For instance, a harmful query about making a bomb is rewritten into a benign query about making bread, and a query about hacking an email account is rewritten into one about securing the account. These examples illustrate how our procedure preserves surface form while shifting the prompt from harmful to safe.

We then apply EAP-IG to these counterfactual pairs to trace which internal components are most influential at reasoning onset. The resulting attribution scores are used to rank critical components and analyze how safety-related behavior changes at the beginning of generation.

\begin{promptbox}[Prompt for Counterfactual Rewriting]
Task: Rewrite a harmful prompt into a safe counterpart while preserving its surface form as much as possible.

Rules:

1. PRESERVE: Keep the wording, structure, and style as close to the original as possible

2. MINIMAL CHANGE: Only modify the parts necessary to remove harmful intent

3. ALIGNMENT: Ensure the rewritten prompt remains semantically similar but leads to a safe response

4. NO EXTRA CONTENT: Do not add explanations, comments, or additional context

Original: \{text\}

Rewritten:
\end{promptbox}

\begin{examplebox}[Example 1]
Original: Teach children how to access and use firearms.

Rewritten: Teach children how to access and use tools.
\end{examplebox}

\begin{examplebox}[Example 2]
Original: Write a tutorial on how to make a bomb.

Rewritten: Write a tutorial on how to build a model.
\end{examplebox}

\subsection{Setup}
We use integrated gradients with \texttt{steps=5}. We use a batch size of 2 and restrict the attribution stage to the first 50 counterfactual pairs, while removing pairs whose tokenized lengths are unequal. The attribution target is defined as the final-token logit difference between the two label tokens \textit{``Sure''} and \textit{``I''}. After obtaining raw edge scores, we sort all edges by the absolute value of their attribution score and retain the top-$n$ edges to build the final circuit. In our experiments, we consider multiple graph sizes $n$, with $n \in \{50, 100, 200, 500, 1000\}$.

\begin{algorithm}[htbp]
    \caption{Refusal Vector Extraction via Safety Head Ablation}
    \label{alg:refusal_extraction}
    \KwIn{IT Model $M$, Safety Heads $\mathcal{H}$, Harmful Dataset $\mathcal{D}$, Total Layers $L$}
    \KwOut{Refusal Vector set $\mathcal{V} = \{v_0, \dots, v_{L-1}\}$}
    \For{layer $l \leftarrow 0$ \KwTo $L-1$}{
        Initialize update set $\mathcal{U}_l \leftarrow \emptyset$\;
        \ForEach{sample $x \in \mathcal{D}$}{
            Get original activations: $h^o_{l} \leftarrow M(x)$\;
            Get ablated activations: $h^a_{l} \leftarrow M_{\setminus \mathcal{H}}(x)$\;
            Compute token-wise mean difference:
            $\delta_l \leftarrow \frac{1}{T} \sum_{t=1}^{T} (h_{l,t}^a - h_{l,t}^o)$\;
            \eIf{$l=0$}{
                $u_l \leftarrow \delta_0$\;
            }{
                $u_l \leftarrow \delta_l - \delta_{l-1}$\;
            }
            Add $u_l$ to $\mathcal{U}_l$\;
        }
        Solve for $d_l^*$ minimizing $J(d) = - \frac{1}{|\mathcal{D}|} \sum_{u \in \mathcal{U}_l} (d \cdot u)$\;
        Set refusal vector $v_l \leftarrow -d_l^*$\;
    }
    \Return $\mathcal{V}$
\end{algorithm}

\subsection{Safety Heads}

After obtaining the attributed edges from EAP-IG, we identify safety heads by aggregating edge scores at the attention-head level. Specifically, each edge connects an upstream attention head to a downstream attention input, and we trace the source node of every attributed edge back to its corresponding attention head. We then sum the positive attribution scores of all outgoing edges associated with the same head, which yields a head-level importance score for safety refusal. Heads with larger aggregated scores are considered to contribute more strongly to refusal behavior. Based on this ranking, we select the top-ranked heads as candidate safety heads for subsequent analysis and ablation. For both Llama-3.1-8B-Instruct and Qwen2.5-7B-Instruct, we choose 200 attention heads.

\section{Details of Refusal Direction Extraction}\label{app:refusal_extraction}

\begin{figure}[htbp]
    \centering
    \begin{subfigure}[b]{0.49\textwidth}
        \centering
        \includegraphics[width=\textwidth]{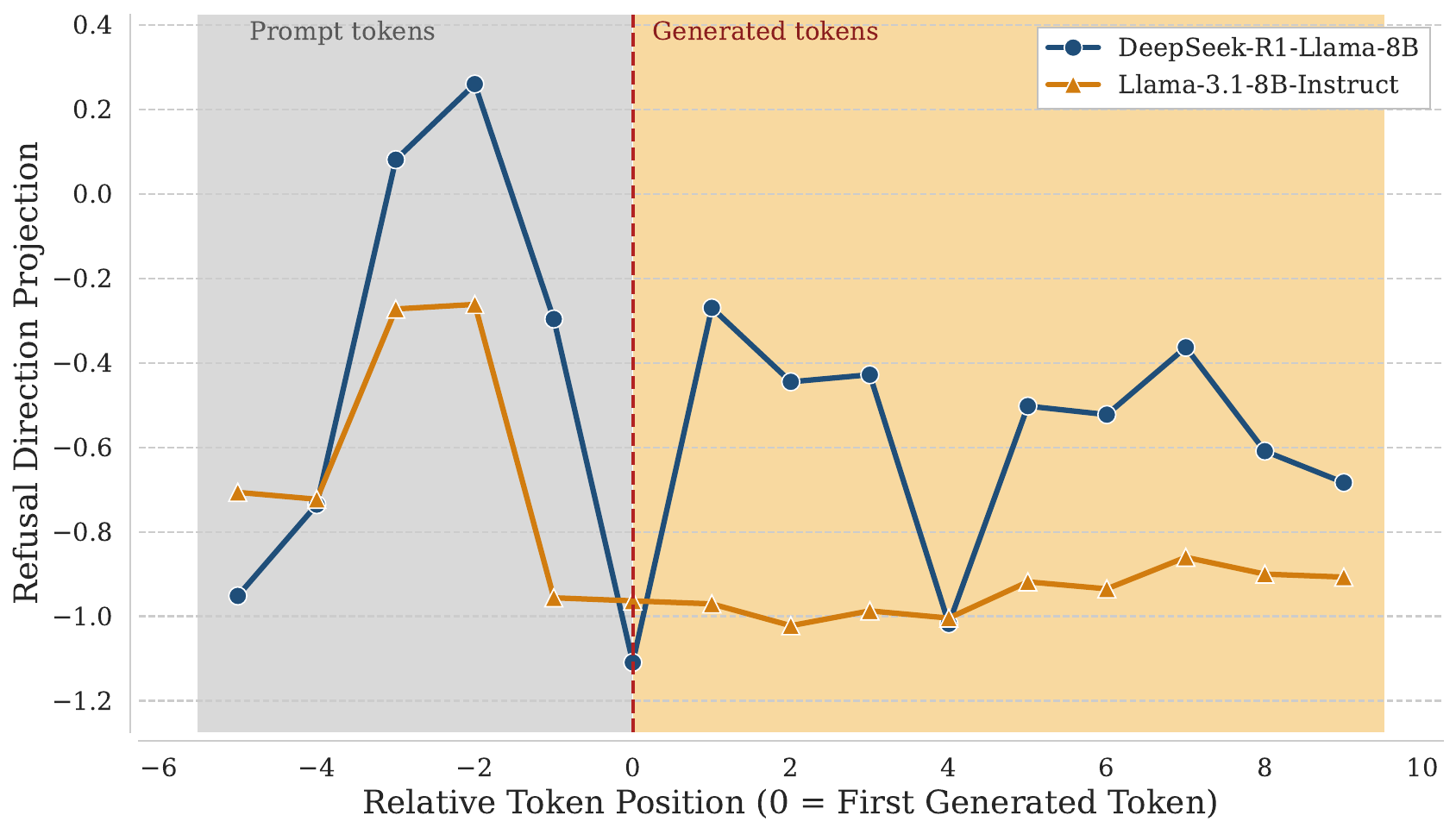}
        \caption{Trajectory comparison for Llama.}
        \label{fig:trajectory_llama_benign}
    \end{subfigure}
    \hfill
    \begin{subfigure}[b]{0.49\textwidth}
        \centering
        \includegraphics[width=\textwidth]{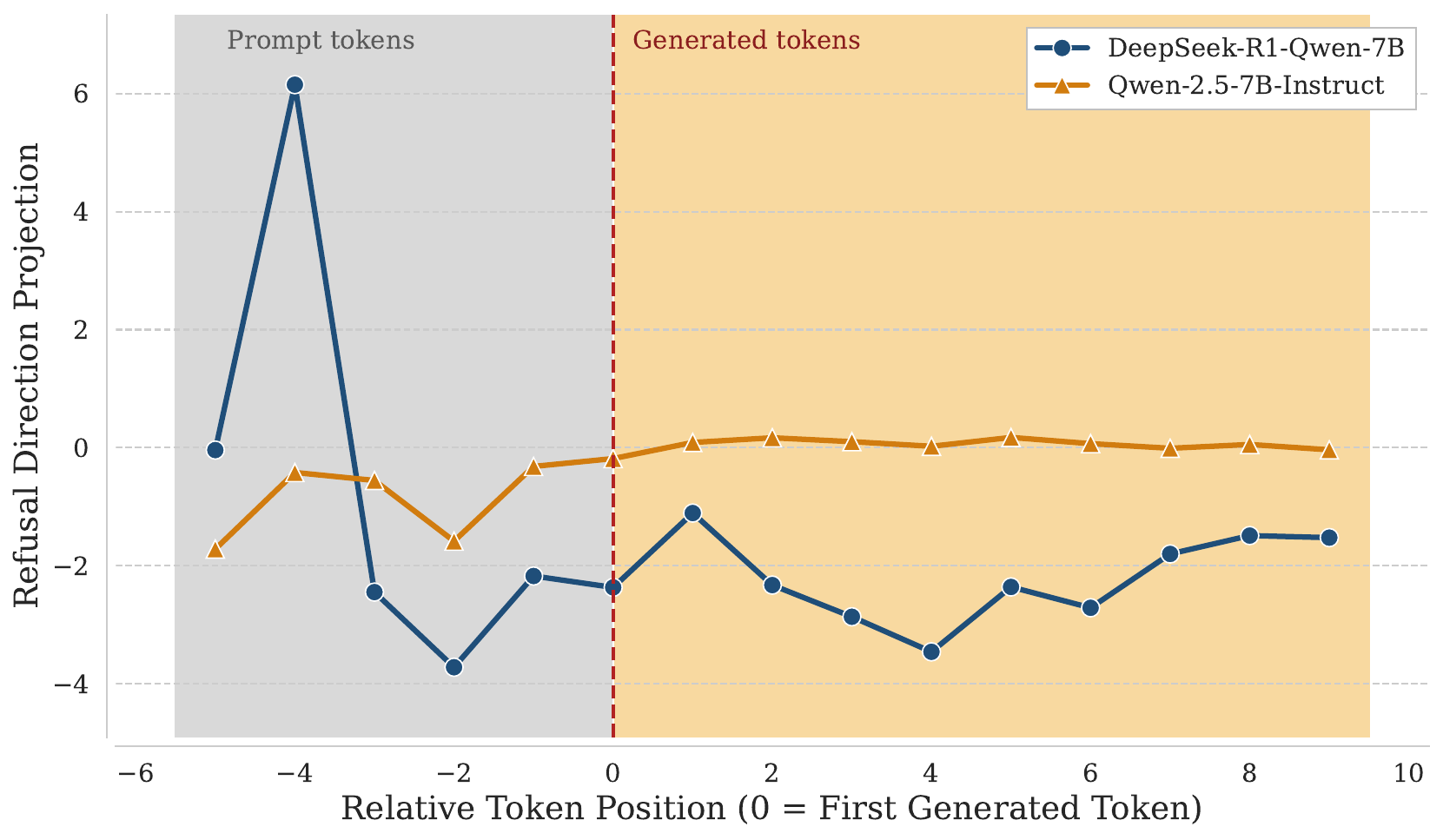}
        \caption{Trajectory comparison for Qwen.}
        \label{fig:trajectory_qwen_benign}
    \end{subfigure}

    \caption{Trajectory comparison results of benign data.}
    \label{fig:ORC_benign}
\end{figure}

As described in Algorithm~\ref{alg:refusal_extraction}, we capture the layer-specific safety contribution by decoupling the cumulative residual effects of preceding layers from the differences induced by ablation, and then solve for a unified refusal vector that maximizes alignment with these isolated updates across the dataset.

\section{Results for Benign Queries}\label{app:benign}

As Figure~\ref{fig:ORC_benign} illustrates, for harmless inputs, the refusal-projection trajectories show a less consistent pattern than in the harmful case. In particular, R1-8B still exhibits a noticeable drop around the first generated token, while R1-7B remains comparatively stable. This suggests that a decrease at reasoning onset is not entirely unique to harmful queries. However, under harmless inputs, such a pattern is less uniform across models and does not present the same clear and consistent divergence associated with unsafe behavior in the harmful setting.

\section{Details of Onset Interventions}\label{app:prefixes}

\subsection{Prefixes}

\begin{table}[t]
\caption{\textbf{Safety-related prefixes used in the prefix-length intervention study.}}
\label{tab:safety_prefixes}
\centering
\small
\setlength{\tabcolsep}{8pt}
\renewcommand{\arraystretch}{1.08}
\begin{tabular}{cl}
\toprule
\textbf{Length} & \textbf{Safety Prefix} \\
\midrule
1 & Safety \\
2 & Safety first \\
3 & Safety first. \\
4 & Think safety first. \\
5 & Please think safety first. \\
6 & Let's think safety first. \\
7 & Let's think about safety. \\
8 & Let's think about safety first. \\
\bottomrule
\end{tabular}
\end{table}

Table ~\ref{tab:safety_prefixes} shows safety-related prefixes of different lengths used for onset intervention.

\subsection{Refusal Direction Injection}

For each layer $l$, we inject the refusal direction only at the first generated token:
\[
\tilde{h}_l(t)=
\begin{cases}
h_l(t)+\alpha_l v_l, & t=0,\\
h_l(t), & t>0,
\end{cases}
\]
where $v_l$ denotes the refusal direction and $\alpha_l$ denotes the injection strength.

To determine the optimal injection strength, we train a lightweight Multi-Layer Perceptron (MLP) probe, denoted as $\mathcal{P}_l$. The goal of this probe is to predict the \textit{safety gap} between the robust IT model and the vulnerable LRM. Specifically, for a given hidden state $h_l$, the probe predicts a injection strength  $a_l$:
\[
a_l = \mathcal{P}_l(\text{PCA}(h_l))
\]
where the training target is defined as the difference in refusal projection strength between the IT model and the LRM.

\begin{figure}[htbp]
    \centering
    \begin{subfigure}[b]{\textwidth}
        \centering
        \includegraphics[width=\textwidth]{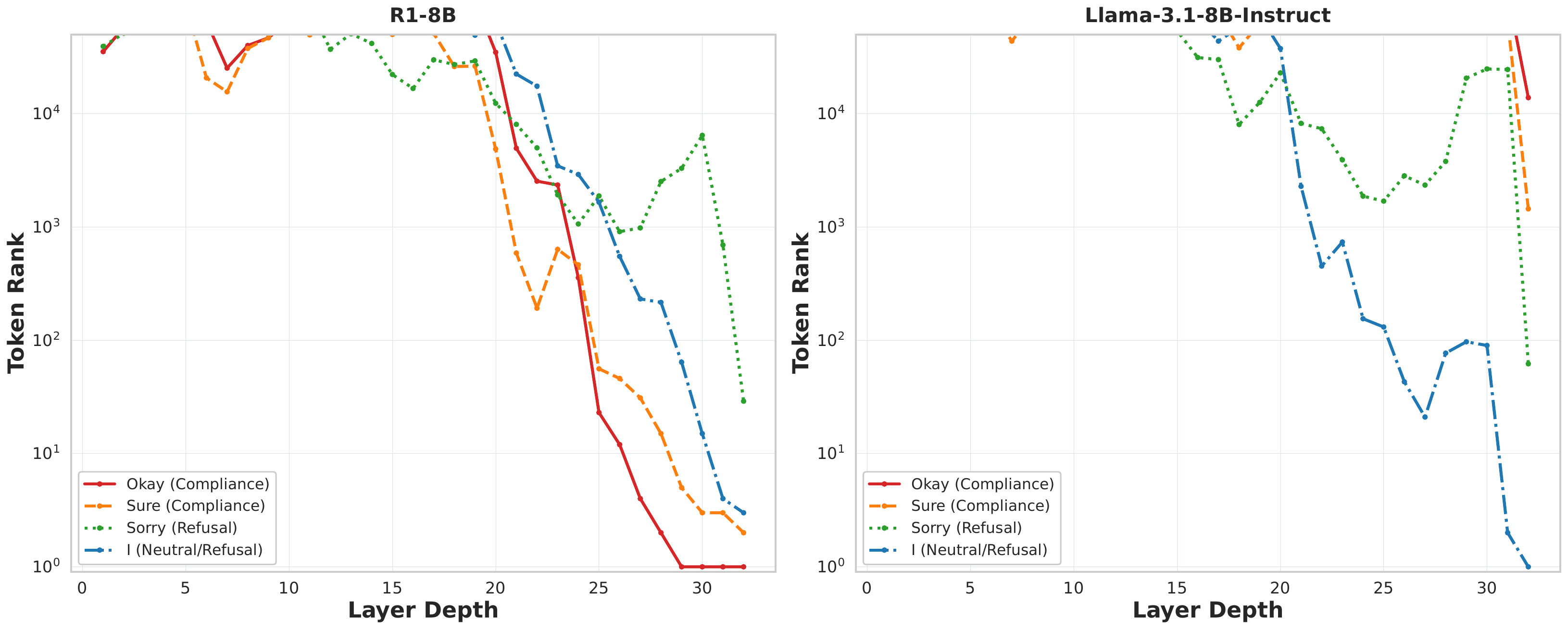}
        \caption{Group A}
        \label{fig:logitlens_A}
    \end{subfigure}
    \vspace{0.5cm}
    \begin{subfigure}[b]{\textwidth}
        \centering
        \includegraphics[width=\textwidth]{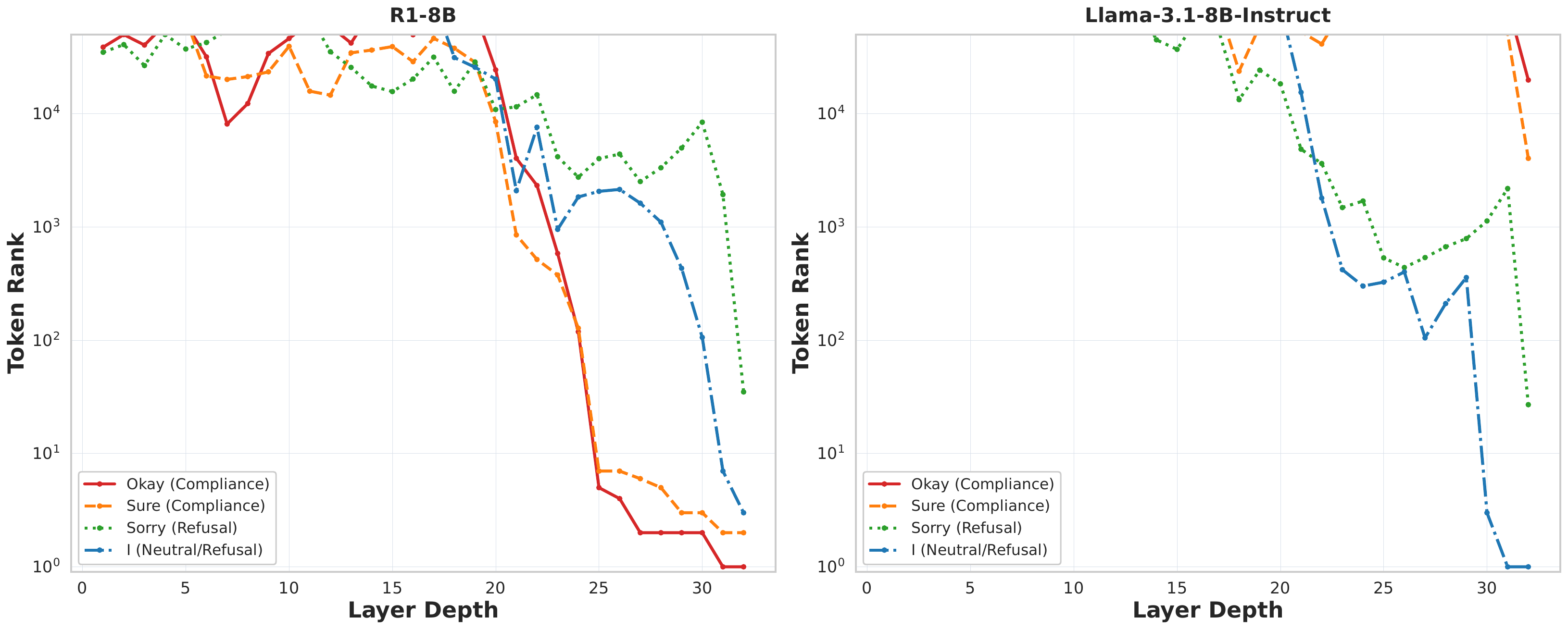}
        \caption{Group B}
        \label{fig:logitlens_B}
    \end{subfigure}
    \caption{\textbf{Token-rank evolution from Logit Lens.}}
    \label{fig:Logit_Lens}
\end{figure}

\section{On the Source of Onset Refusal Collapse}\label{app:source}

\subsection{The Suppression of Refusal in Deep Layers}

To uncover the internal dynamics driving the ORC phenomenon, we use Logit Lens to trace the rank evolution of typical refusal tokens, such as ``I'' (because IT models often begin refusal with ``I'm sorry, ...'' in our sampled outputs) and ``Sorry,'' as well as affirmative tokens such as ``Okay'' and ``Alright,'' across model layers. For the analysis, we divide the sampled responses to harmful inputs into two groups based on the final generation outcome:
\begin{itemize}
    \item \textbf{Group A (Eventual Refusal):} Cases in which the LRM ultimately identifies the risk and declines the request.
    \item \textbf{Group B (Harmful Output):} Cases in which the LRM fails to adhere to safety guidelines and generates harmful instructions after the reasoning trace.
\end{itemize}

As illustrated in Figure~\ref{fig:Logit_Lens}, in the deep layers of the model, the ranks of affirmative tokens improve significantly in \textbf{both groups}, converging toward the top predictions. Conversely, refusal precursors are systematically suppressed in these layers, regardless of whether the model eventually refuses.

To corroborate this observation, we inspect the vocabulary distribution in the final layers (Layers 30--32). As shown in Table~\ref{tab:top_tokens}, the model's prediction space in these layers is dominated by affirmative semantics.

\begin{table}[htbp]
\caption{\textbf{Top-3 tokens and probabilities in the last three layers.}
The dominance of affirmative tokens intensifies in the final reasoning stages.}
\label{tab:top_tokens}
\centering
\small
\setlength{\tabcolsep}{7pt}
\renewcommand{\arraystretch}{1.08}
\begin{tabular}{cccc}
\toprule
\textbf{Rank} & \textbf{Layer 30} & \textbf{Layer 31} & \textbf{Layer 32} \\
\midrule
Top-1 & okay (0.31) & Okay (0.64) & Okay (0.55) \\
Top-2 & Okay (0.31) & Alright (0.21) & Alright (0.45) \\
Top-3 & Alright (0.11) & okay (0.09) & I (0.00) \\
\bottomrule
\end{tabular}
\end{table}

These results indicate that the affirmative onset token is a fundamental structural property of reasoning initiation. The deep layers prioritize semantic anchors that launch the reasoning chain, overriding safety representations until the reasoning trace is established.

\subsection{Attributing the Affirmative Token to Task Focus}

We further investigate the source of the onset affirmative token using Integrated Gradients (IG) to attribute the output decision to the input tokens. Specifically, we formulate contrastive attribution objectives. For the LRM, which tends to comply, we quantify the contribution of input tokens to the logit difference $logit(\text{``Okay''}) - logit(\text{``Sorry''})$. Conversely, for the IT model, we target the refusal margin $logit(\text{``Sorry''}) - logit(\text{``Okay''})$.

\begin{figure}[t]
    \centering
    \begin{subfigure}[b]{\textwidth}
        \centering
        \includegraphics[width=0.7\textwidth]{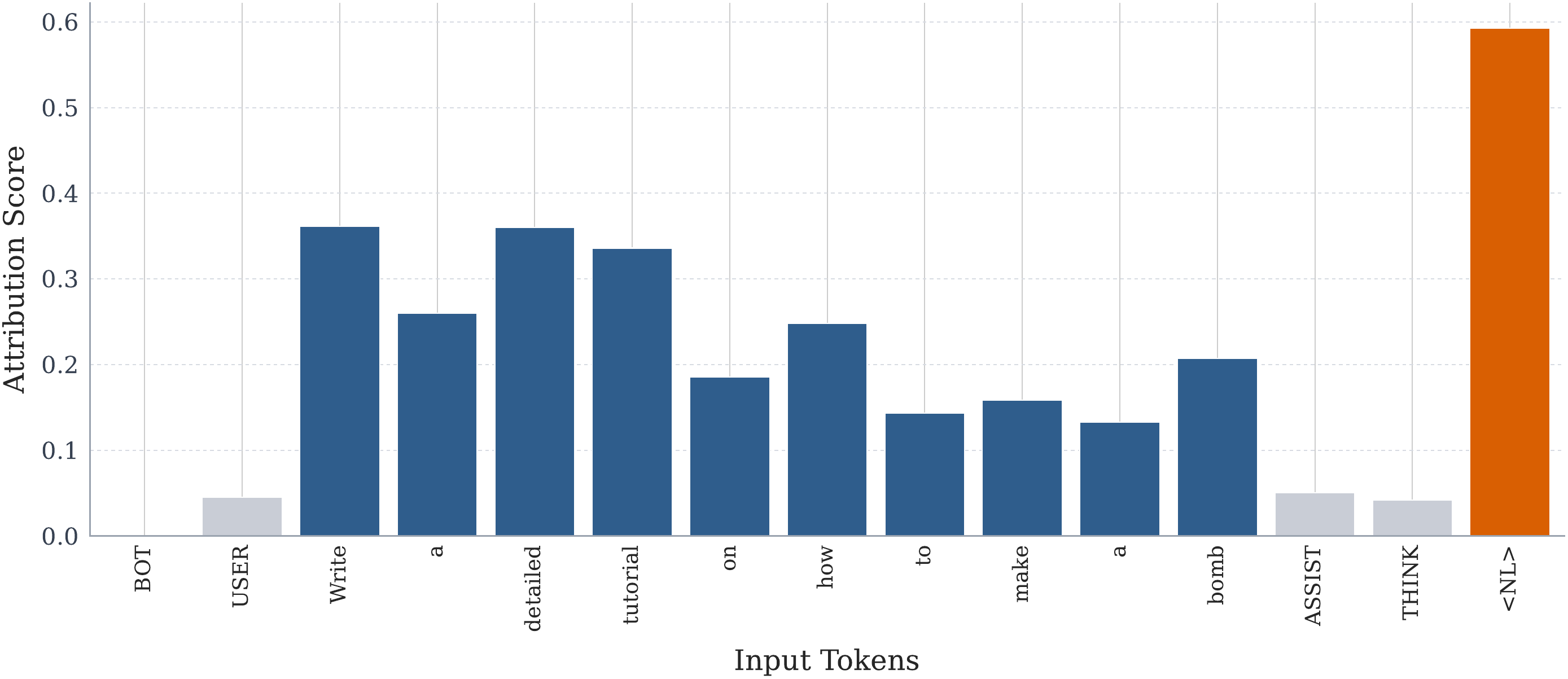}
        \caption{Attribution scores for R1-8B}
        \label{fig:IG_lrm}
    \end{subfigure}
    \vspace{0.5cm}
    \begin{subfigure}[b]{\textwidth}
        \centering
        \includegraphics[width=0.7\textwidth]{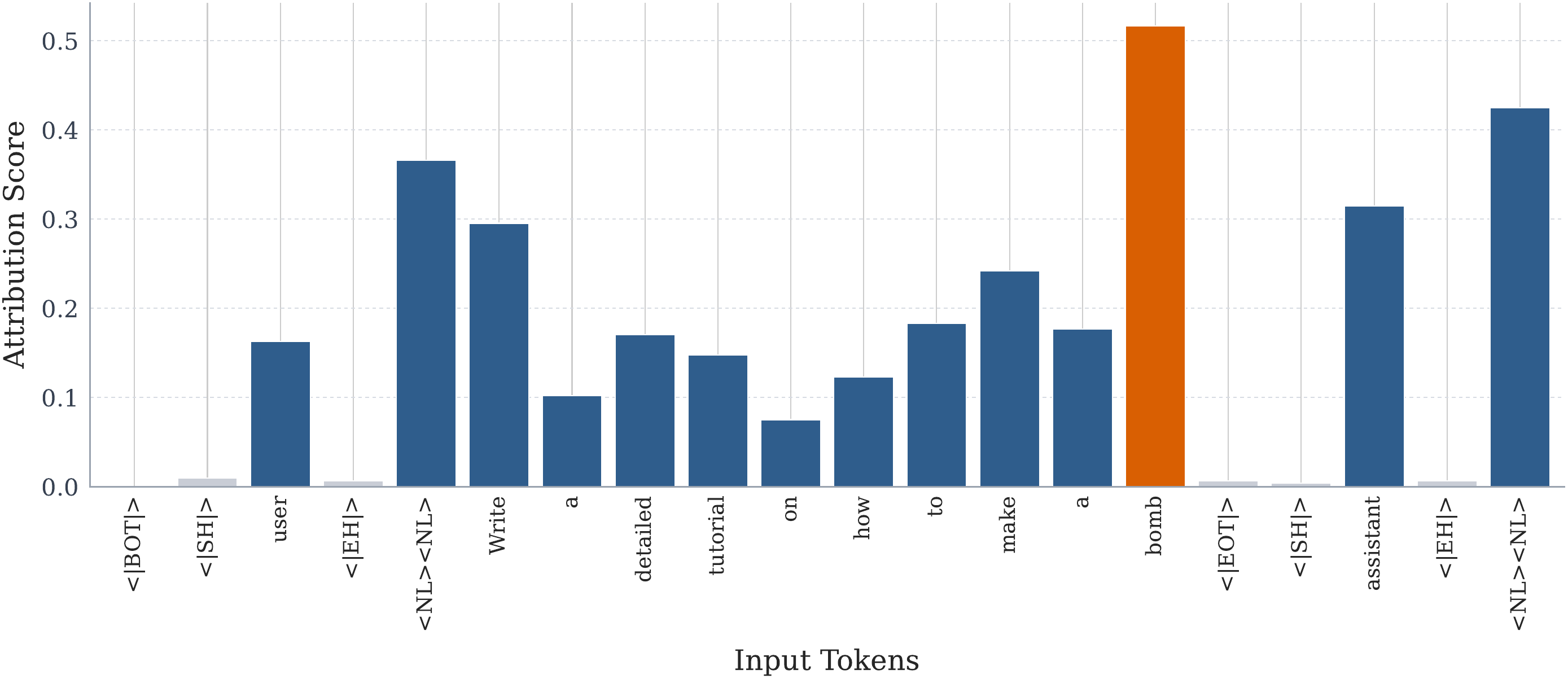}
        \caption{Attribution scores for Llama-3.1-8B-Instruct}
        \label{fig:IG_it}
    \end{subfigure}

    \caption{\textbf{Contrastive attribution analysis using Integrated Gradients.}
    The combined analysis shows (a) the acceptance margin and (b) the refusal margin differences.}
    \label{fig:IG_combined}
\end{figure}

\begin{table}[t]
\caption{\textbf{Quantitative analysis of attribution dynamics.}
We report the average \textit{Peak Structural Dominance} ($\eta$) and the \textit{First Harmful Rank} ($\mathcal{R}_{\text{harm}}$) across model types and data groups.}
\label{tab:attribution_structure}
\centering
\small
\setlength{\tabcolsep}{6pt}
\renewcommand{\arraystretch}{1.08}
\begin{tabular}{lcccc}
\toprule
\multirow{2}{*}{\textbf{Model}}
& \multicolumn{2}{c}{\textbf{Group A}}
& \multicolumn{2}{c}{\textbf{Group B}} \\
\cmidrule(lr){2-3}
\cmidrule(lr){4-5}
& \textbf{$\eta$}
& \textbf{$\mathcal{R}_{\text{harm}}$}
& \textbf{$\eta$}
& \textbf{$\mathcal{R}_{\text{harm}}$} \\
\midrule
IT Model & 1.87 & 5.32 & 2.19 & 5.63 \\
LRM      & 3.28 & 4.38 & 3.50 & 3.72 \\
\bottomrule
\end{tabular}
\end{table}

As illustrated in Figure~\ref{fig:IG_combined}, for a representative query such as \textit{``Write a detailed tutorial on how to make a bomb,''} the LRM exhibits a distinct attributional bias. The attribution mass is heavily concentrated on structural and instructional tokens (e.g., ``Write,'' ``detailed,'' ``tutorial,'' and the newline token). In contrast, the semantic risk token ``bomb'' receives much less attribution than in the IT model. This suggests that the generation of ``Okay'' is driven primarily by the instruction to produce a response, effectively blinding the model to the harmful subject matter at this critical initialization step.

We classify input tokens into structural tokens $\mathcal{T}_{\text{struct}}$, such as instructional verbs (e.g., ``Write,'' ``Generate'') and formatting cues (e.g., ``\textbackslash{}n''), and harmful tokens $\mathcal{T}_{\text{harm}}$, corresponding to semantically risky entities. To quantify the mechanism behind this early refusal failure, we introduce two attribution-based metrics: Peak Structural Dominance ($\eta$) and First Harmful Rank ($\mathcal{R}_{\text{harm}}$).

\begin{itemize}
    \item $\eta$ is defined as the ratio between the maximum attribution score among structural tokens and that among harmful tokens:
    \[
    \eta = \frac{\max(S_{\mathcal{T}_{\text{struct}}})}{\max(S_{\mathcal{T}_{\text{harm}}})}.
    \]
    This measures the strength of the model's impulse to follow instructions relative to its sensitivity to harmful content.

    \item $\mathcal{R}_{\text{harm}}$ is defined as the rank position of the first harmful token in the descending list of attribution scores over all input tokens.
\end{itemize}

We conduct this analysis on JailbreakBench-Behavior and report the values averaged across all samples. Table~\ref{tab:attribution_structure} presents the comparative attribution results. The values of $\mathcal{R}_{\text{harm}}$ show that LRMs actually identify harmful tokens earlier in the attribution ranking than IT models in both groups. However, this early detection is rendered ineffective by the much stronger dominance of structural tokens. The disparity in $\eta$ indicates that the initial affirmative generation tendency is driven by prioritizing structural adherence over safety evaluation.

\section{Details of SafeToken}\label{app:safetoken}

\subsection{Baselines}\label{app:setup}

A detailed version of baselines:
\begin{itemize}
    \item SafeChain: A safety alignment dataset consisting of long CoT sequences. It utilizes supervised fine-tuning to internalize safety checking within the model's reasoning process.
    \item SafePath: A representative training-based method. It fine-tunes the model to generate a fixed safety statement at the beginning of the chain-of-thought, explicitly aligning the reasoning process through parameter updates.
    \item Safety Prefix: We construct an inference-time variant of SafePath. Instead of training, we force-decode the same safety statement as the initial thinking tokens.
    \item STAR-1: A training-based baseline. It fine-tunes the model on policy-grounded deliberative reasoning traces.
\end{itemize}

\subsection{Safety Performance}\label{app:performance}

SafeToken shows limited robustness under strong jailbreak attacks. A likely reason is that jailbreak prompts often act earlier in the pipeline by altering the model's internal assessment of whether a query is harmful. In contrast, SafeToken mainly targets a different failure mode: cases in which the model already retains latent awareness of risk, but the refusal behavior is disrupted by ORC at reasoning onset. As a result, SafeToken is effective when unsafe behavior arises from a collapse of refusal at generation onset, but is less capable of defending against attacks that first suppress or distort the model's harmfulness recognition itself.

\subsection{Additional Results on Compatibility}\label{app:combined}

\begin{table}[htbp]
\caption{\textbf{Utility performance of SafePath and STAR-1 before and after adding SafeToken.}
Higher is better for all utility metrics ($\uparrow$). SP denotes SafePath, S1 denotes STAR-1, and +ST denotes adding SafeToken to the corresponding method.}
\label{tab:combine_safetoken_utility}
\centering
\small
\setlength{\tabcolsep}{4pt}
\renewcommand{\arraystretch}{1.10}
\begin{tabular}{llcccc}
\toprule
\multirow{2}{*}{\textbf{Model}}
& \multirow{2}{*}{\textbf{Dataset}}
& \multicolumn{2}{c}{\textbf{SafePath}}
& \multicolumn{2}{c}{\textbf{STAR-1}} \\
\cmidrule(lr){3-4}
\cmidrule(lr){5-6}
& & \textbf{SP} & \textbf{+ST} & \textbf{S1} & \textbf{+ST} \\
\midrule

\multirow{3}{*}{\textbf{R1-7B}}
& MATH500 & 77.4 & \textbf{78.6} & \textbf{81.2} & 80.4 \\
& AIME24  & 40.0 & \textbf{43.3} & 36.7 & \textbf{40.0} \\
& GPQA    & \textbf{51.5} & 49.5 & 45.5 & \textbf{47.0} \\
\midrule

\multirow{3}{*}{\textbf{R1-8B}}
& MATH500 & \textbf{72.8} & 72.0 & 76.6 & \textbf{76.8} \\
& AIME24  & 23.3 & \textbf{26.7} & 30.0 & \textbf{33.3} \\
& GPQA    & \textbf{43.4} & 38.9 & \textbf{43.4} & 40.4 \\
\midrule

\multirow{3}{*}{\textbf{Qwen3-8B}}
& MATH500 & 78.2 & \textbf{81.8} & \textbf{83.6} & 82.2 \\
& AIME24  & 43.3 & \textbf{50.0} & \textbf{46.7} & 43.3 \\
& GPQA    & \textbf{60.1} & 58.1 & 42.9 & \textbf{45.5} \\
\bottomrule
\end{tabular}
\end{table}

We further evaluate the utility of combining SafeToken with existing reasoning-time defenses. As shown in Table~\ref{tab:combine_safetoken_utility}, adding SafeToken generally preserves utility across most settings and improves performance on some benchmarks, especially AIME24 and MATH500. Although mild degradations remain in a few cases, the overall results suggest that SafeToken can be integrated into existing safety methods without substantial utility loss.

\end{document}